\documentclass[conference,hidelinks]{IEEEtran}
\usepackage{cite}
\usepackage{amsmath,amssymb,amsfonts}
\usepackage{algorithm2e}
\usepackage{graphicx}
\usepackage{textcomp}
\usepackage{xcolor}
\usepackage{url}
\usepackage{subcaption}
\usepackage{gensymb}
\usepackage{textcomp}
\usepackage{enumitem}

\DeclareMathOperator*{\argminA}{arg\,min}

\IEEEoverridecommandlockouts

\newcommand{\EWA}{\mathbin{\text{$\vcenter{\hbox{\textcircled{$+$}}}$}}}

\def\BibTeX{{\rm B\kern-.05em{\sc i\kern-.025em b}\kern-.08em
    T\kern-.1667em\lower.7ex\hbox{E}\kern-.125emX}}
\begin{document}

\title{ConvTimeNet: A Pre-trained Deep Convolutional Neural Network for Time Series Classification\\}

\author{\IEEEauthorblockN{Kathan Kashiparekh\textsuperscript{*}\thanks{\textsuperscript{*}Work done during internship at TCS Research. Extended version of paper accepted at IJCNN'19.}}
	\IEEEauthorblockA{
		BITS-Pilani Goa Campus, Goa\\
		India\\
		Email: f20140792@goa.bits-pilani.ac.in}
	\and 
	\IEEEauthorblockN{Jyoti Narwariya, Pankaj Malhotra, Lovekesh Vig and
		Gautam Shroff}
	\IEEEauthorblockA{TCS Research, New Delhi\\
		India\\
		Email: \{jyoti.narwariya,malhotra.pankaj,lovekesh.vig,gautam.shroff\}@tcs.com}}

\maketitle

\begin{abstract}
Training deep neural networks often requires careful hyper-parameter tuning and significant computational resources. 
In this paper, we propose ConvTimeNet (CTN): an off-the-shelf deep convolutional neural network (CNN) trained on diverse univariate time series classification (TSC) source tasks. 
Once trained, CTN can be easily adapted to new TSC target tasks via a small amount of fine-tuning using labeled instances from the target tasks.
We note that the length of convolutional filters is a key aspect when building a pre-trained model that can generalize to time series of different lengths across datasets.
To achieve this, we incorporate filters of multiple lengths in all convolutional layers of CTN to capture temporal features at multiple time scales. 
We consider all 65 datasets with time series of lengths up to 512 points from the UCR TSC Benchmark for training and testing transferability of CTN:  
We train CTN on a randomly chosen subset of 24 datasets using a multi-head approach with a different softmax layer for each training dataset, and study generalizability and transferability of the learned filters on the remaining 41 TSC datasets. 
We observe significant gains in classification accuracy as well as computational efficiency when using pre-trained CTN as a starting point for subsequent task-specific fine-tuning compared to existing state-of-the-art TSC approaches.
We also provide qualitative insights into the working of CTN by: i) analyzing the activations and filters of first convolution layer suggesting the filters in CTN are generically useful, ii) analyzing the impact of the design decision to incorporate multiple length decisions, and iii) finding regions of time series that affect the final classification decision via occlusion sensitivity analysis. 

\end{abstract}


\section{Introduction}
In the current digital era, time series data is ubiquitous due to widespread adoption of Internet of Things technology with applications across several domains such as healthcare, equipment health monitoring, meteorology, demand forecasting, etc.
Time series classification (TSC) has several practical applications such as those in healthcare (e.g. real-time monitoring, disease diagnosis using time series of physiological parameters, classifying heart arrhythmia in ECG, etc.) and fault diagnostics using sensor data from equipment (e.g. determining the type of fault from sensor data).

Deep learning approaches, such as those based on recurrent neural networks (RNNs) and convolution neural networks (CNNs) have been proven to be very effective for univariate time series classification (UTSC) \cite{wang2017time,karim2018lstm,fawaz2018deep}.
Deep CNNs have yielded some of the state-of-the-art models \cite{wang2017time,fawaz2018deep} for TSC.
However, it is well-known that training deep networks requires significant hyper-parameter tuning effort and expertise, high computational resources, and is prone to overfitting, especially when access to a large labeled training dataset is difficult.

Transfer learning \cite{pan2010survey,bengio2012deep} is known to be an effective way to address some of the above-mentioned challenges in training deep neural networks: It enables knowledge transfer from a \textit{source} task with sufficient training instances to a related \textit{target} task with fewer training instances, for example, by training a deep neural network model on source task(s) with large labeled data, and adapting this model for the target task using a small amount of labeled data from the target task.
This approach of fine-tuning a pre-trained network for target task is often faster and easier than obtaining a network from scratch (starting with randomly initialized weights) that often requires computationally expensive hyper-parameter tuning \cite{bengio2012deep}.
For example, it is well-established that training a deep CNN using a diverse set of images results in generic filters that can provide useful features for target tasks with images from unseen domains \cite{simonyan2014very}.

Recently, transfer learning for TSC using deep neural networks has been explored, e.g., using RNNs in \cite{malhotra2017timenet,gupta2018using,gupta2018transfer}, and using CNNs in \cite{serra2018towards,fawaz2018transfer}. 
These approaches pre-train a deep network on time series from diverse domains, and then either use it as a time series feature extractor for the target task as in TimeNet \cite{malhotra2017timenet,gupta2018using}, 
or use the pre-trained network to initialize the parameters of the neural network for the target task \cite{serra2018towards,fawaz2018transfer,gupta2018transfer}.
When pre-training a deep network on time series from diverse domains, the rate of change of relevant information in time series can vary significantly across tasks and domains. 
We note that CNN based TSC architectures, as proposed in \cite{wang2017time,fawaz2018deep}, can extract local information at only one time scale determined by a single fixed filter size, limiting the flexibility of the model.
The filter size of a convolutional layer should, therefore, be chosen carefully to extract relevant features depending on the domain and target task.
Indeed, hand-crafted transformations such as smoothening and down-sampling of time series to learn features at various time scales for TSC have been shown to be useful in Multi-scale CNNs \cite{cui2016multi}.
We hypothesize that this aspect is even more relevant in a transfer learning setting when adapting a pre-trained deep CNN with its convolutional filters to a target domain.
For example, training a common network on diverse tasks with time series as short as $\approx 20$ to as long as $\approx 2000$ warrants the need to take into account varying relevant time scales: a filter length of $5$ may be useful and sufficient to capture relevant features for datasets with short time series, whereas a filter length of $50$ may be more appropriate for datasets with long time series.

In this work, we propose \textit{ConvTimeNet} (CTN), a deep CNN-based transfer learning approach for UTSC. CTN consists of multiple length 1-D convolutional filters in all convolutional layers (similar to that in InceptionNet \cite{szegedy2015going,roy2018chrononet}) resulting in filters that can capture features at multiple time scales. 
The key contributions of this work can be summarized as follows:
\begin{enumerate}
	\item We propose ConvTimeNet (CTN), a novel pre-trained deep CNN for univariate time series classification tasks.
	\item We demonstrate that fine-tuning CTN for target tasks outperforms a deep CNN trained from scratch in terms of classification accuracy. 
	\item We demonstrate that the CTN can cater to diverse time series of varying lengths by using 1-D convolutional filters of multiple lengths to capture features at different time scales.
	\item We demonstrate that fine-tuning the CTN is computationally efficient compared to training a deep CNN from scratch.
\end{enumerate}
We report state-of-the-art results for UTSC on the UCR TSC benchmark \cite{UCRArchive} considering all datasets with time series length up to 512, while also significantly improving upon our RNN-based TimeNet \cite{timenet}.

\section{Related Work\label{sec:rw}}
Several approaches for UTSC have been reviewed in \cite{bagnall2017great}.
A plethora of research has been done using feature-based approaches or methods to extract a set of features that represent time series patterns as reviewed in \cite{bagnall2017great}.
COTE (Collective of Transformation-Based Ensembles), ST (Shapelet Transform), PF (Proximity Forest) and  (Bag-of-SFA-Symbols) are considered to be the state-of-the-art non-deep learning algorithms for UTSC \cite{bagnall2017great}.
Although COTE is one of the most accurate classifiers, it has a large training time complexity which is $O(N^2 · T^4 )$ ($N$ being number of training samples and $T$ being time series length).
Whereas most of these approaches extract features using data from the UTSC task at hand, our proposed approach aims to learn generic multi-time-scale features via filters in CNNs which can be useful on time series from unseen domains in a transfer learning setting.

Recently, several deep learning architectures based on Long short-term memory networks (LSTMs), CNNs, and their combinations have been proposed for univariate TSC (e.g. \cite{cui2016multi,wang2017time,karim2018lstm}).
To overcome overfitting issues and achieve better generalizability, data augmentation methods have been proposed: combining datasets of similar length across domains \cite{le2016data}, using simulated data \cite{fawaz2018data}, window slicing, warping, and mixing \cite{le2016data,cui2016multi}, etc. 
Decorrelating filters of CNNs has been recently shown to be effective in reduce overfitting \cite{kaushal2019regularizing}.
On the other hand, we consider transfer learning to achieve better generalizability by pre-training a model on large labeled datasets and then fine-tuning it for the end (target) task with potentially less labeled data.

Several approaches for transfer learning exist in other domains such as computer vision and natural language processing, e.g. via fine-tuning \cite{girshick2014rich,long2015learning}.
In the context of time series classification applications, few instances of leveraging transfer learning to achieve better generalizability have been considered when using deep learning models, e.g. \cite{malhotra2017timenet,serra2018towards,fawaz2018transfer}. 
\cite{ukil2019timenet} reports significant improvements by combining pre-specified features from Fourier, wavelet, and other transformations of the time series signals with deep learning features from TimeNet \cite{timenet}.
However, none of these approaches consider the inherent need of multi-scale learning when training a common model across domains with varying time series.
\cite{roy2018chrononet} attempts to address this need but not in transfer learning scenarios.
In this work, we propose CTN which uses multiple length filters yielding significant improvements over fixed-length CNN models for transfer learning.

\section{ConvTimeNet\label{sec:ctn}}

\subsection{Overview\label{sec:overview}}
Consider a univariate time series $x_{1\ldots T} = x_1,x_2,\dots,x_T$ with $x_t \in \mathbb{R}$ for $t=1,\dots,T$, and $T$ being the length of the time series. 
Further, consider a labeled dataset $\mathcal{D}_j=\{(x^{(i)}_{1\ldots T^{(i)}},c^{(i)})\}_{i=1}^{N_j}$ having $N_j$ samples and ground truth class label $c^{(i)} \in \{c_1,\dots,c_{K_j}\}$ with $K_j$ being the number of classes. 
The goal of UTSC model trained on $\mathcal{D}_j$ is to predict a probability vector $\hat{\mathbf{y}} = [\hat{y}_1, \hat{y}_2,\ldots, \hat{y}_{K_j}]$ with $\sum_{k=1}^{K_j}\hat{y}_k=1$ corresponding to the ground truth one-hot vector $\mathbf{y} \in \{0,1\}^{K_j}$ for a test time series $\hat{x}_{1\ldots T}$.
In this work, we propose CTN, a deep convolutional neural network (CNN) based UTSC model that is trained on a \textit{source} set $\mathcal{S} = \{\mathcal{D}_j\}_{j=1}^{S}$ containing $S$ UTSC source datasets. 
Once trained, CTN can be adapted to a new target UTSC task with small labeled dataset $\mathcal{D}'$ via suitable fine-tuning.

We first describe the architecture considered for CTN in Section \ref{ssec:arch}, and then describe how we train CTN using a diverse source set in Section \ref{ssec:training_ctn}. In Section \ref{ssec:fine-tune}, we describe how CTN is adapted for a target task via fine-tuning.

\subsection{CTN Architecture\label{ssec:arch}}
\begin{figure}[h]
	\centering
	\includegraphics[scale=0.35, angle=-90,trim={2cm, 2cm, 8cm, 0cm},clip]{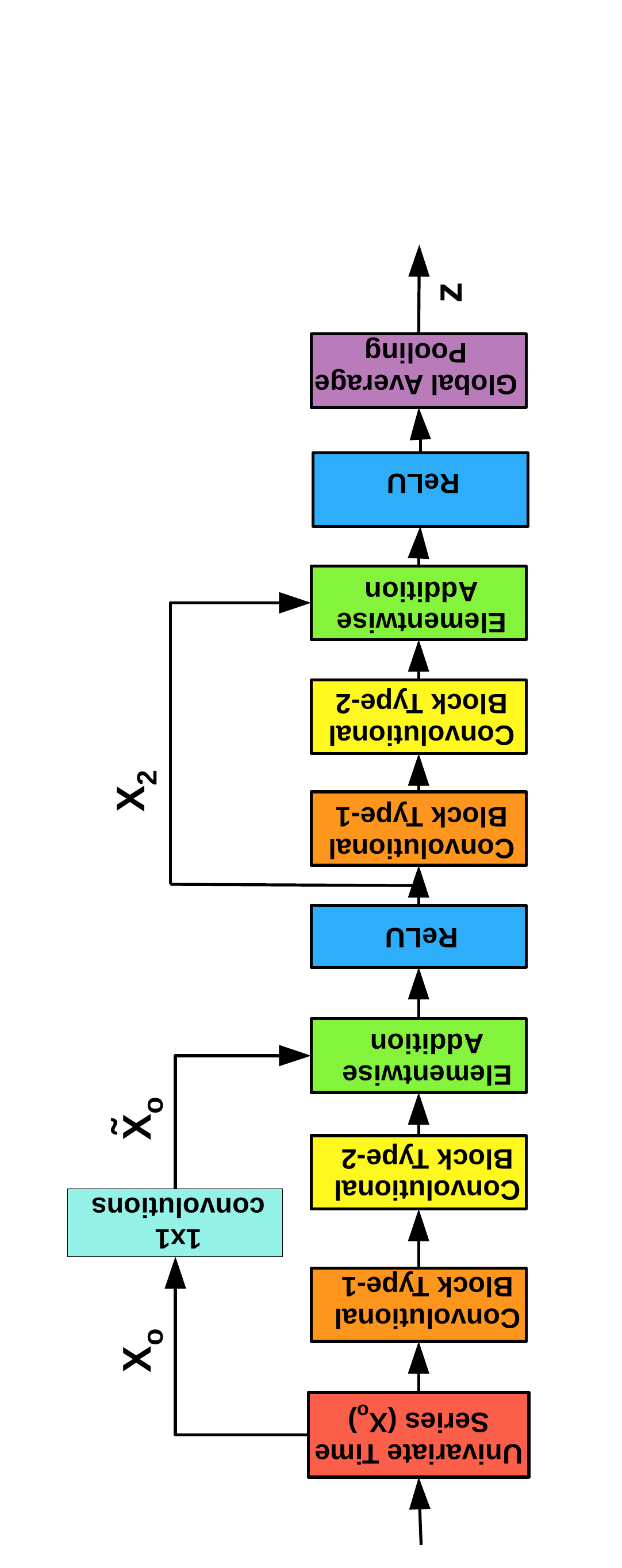}
	\caption{ConvTimeNet with four convolutional blocks.}
	\label{fig:convtimenet}
\end{figure}

As depicted in Fig. \ref{fig:convtimenet}, the architecture of CTN is fairly simple: it consists of multiple convolutional blocks followed by a Global Average Pooling (GAP) layer \cite{lin2013network}, as detailed next.
CTN is trained via additional multi-head fully connected (FC) and softmax layers, as detailed in Section \ref{ssec:training_ctn}. 

\subsubsection{Convolutional Blocks with multiple length filters}
 
\begin{figure}[h]
	\centering
	\includegraphics[scale=0.4,trim={0cm, 13.5cm, 0cm, 2cm},clip]{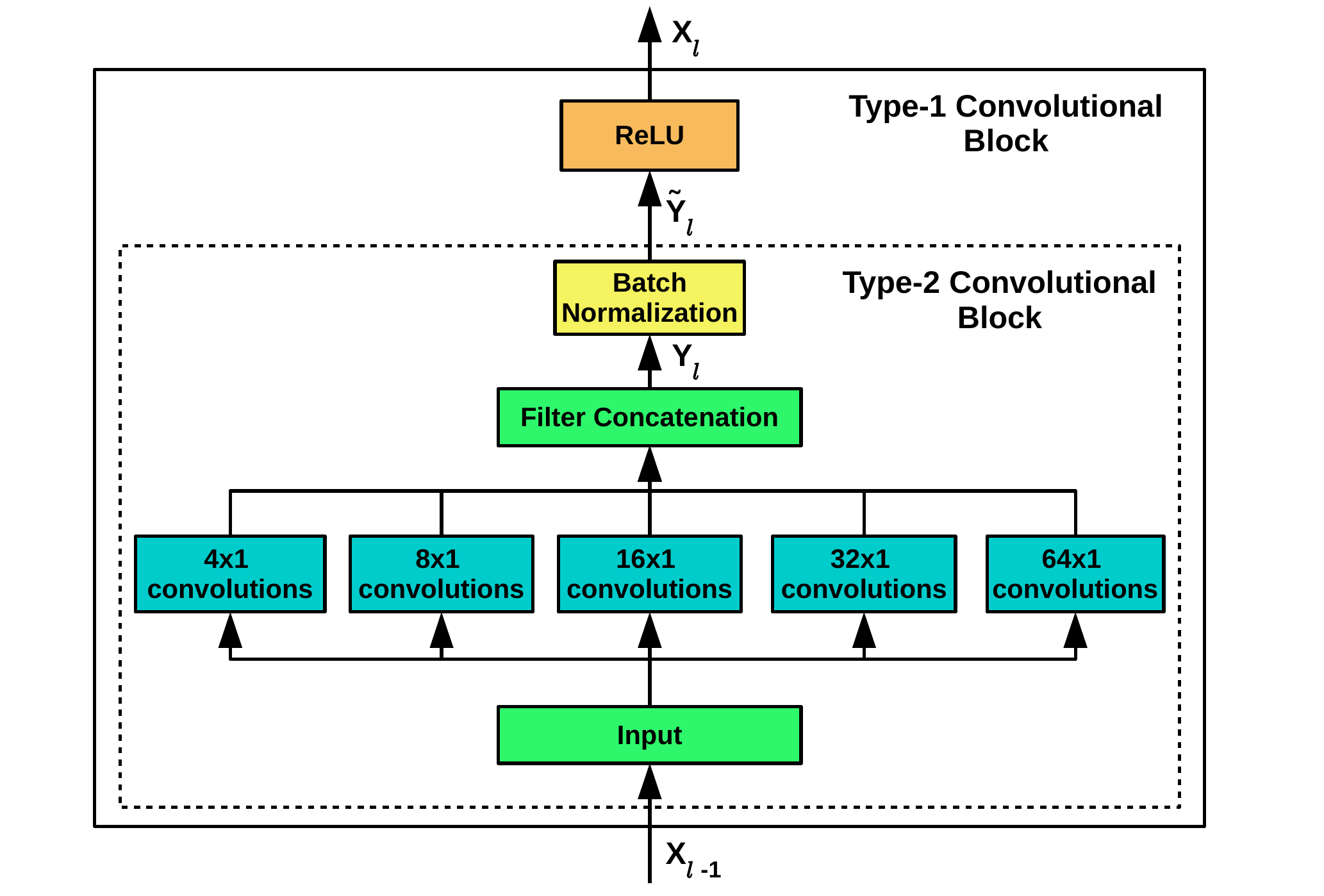} 
	\caption{Convolutional Block}
	\label{fig:conv_block}
\end{figure}

Consider a CTN with $L$ convolutional blocks. The $l$-th convolutional block consists of $m_l$ 1-D convolution filters of varying lengths (e.g. filters of exponentially varying lengths $4,8,16,32,64$) as shown in Fig. \ref{fig:conv_block}. 
This allows CTN to extract and combine features from different time scales, improving the generalization of the network across diverse TSC tasks (as shows later empirically in Section \ref{ssec:ablation}).
 
For a time series $x_{1\ldots T}$, the input tensor to the $l$-th convolutional block with $m_l$ filters is given by $X_{l-1} \in \mathbb{R}^{T \times m_{l-1}}$ with $m_{l-1}$ channels (Note: $m_0=1$ corresponding to the univariate input time series). 
A filter of length $f$ in layer $l$ is represented by a tensor $W_{k,l} \in \mathbb{R}^{f \times m_{l-1}}$ where $k=1,\ldots,m_l$. 
The feature map obtained using $k$-th filter is given by $W_{k,l}*X_{l-1} + b_{k,l}$ where $*$ is the convolution operation and $b_{k,l} \in \mathbb{R}$ is a scalar bias\footnote{We use zero-padding to keep the length $T$ of input and output same.}. 
The output tensor consisting of the feature maps from the $m_l$ filters is represented by $Y_{l} \in \mathbb{R}^{T \times m_l}$.
Note that the length $f$ varies across filters, e.g. $f=4,\ldots,64$ in Fig. \ref{fig:conv_block}. We consider equal number of filters for each length, such that there are $m_l/5$ filters of each length.

We consider residual connections \cite{he2016deep} across blocks to allow gradients to flow directly to lower layers, enabling training of deep networks.
Depending on whether the output of the block is to be added to the output of a previous layer in the network via residual connections or not, there are two types of convolutional blocks: Type-1 and Type-2, as shown in Fig. \ref{fig:conv_block}. 
For convolutional blocks of Type-1, $Y_{l}$ is passed through a batch normalization (BN) layer \cite{ioffe2015batch} and a Rectified Linear Unit (ReLU) layer (where $ReLU(x)=max(0,x)$ operation) to obtain $X_l = ReLU(BN(Y_l))$. 
The structure of the convolutional blocks of Type-2 differs from that of Type-1 in the sense that $Y_l$ is processed by the BN layer to obtain $\tilde{Y}_l = BN(Y_l)$ but not the ReLU layer thereafter. 
Instead, a residual connection is used such that $X_{l-2}\in \mathbb{R}^{T\times m_{l-2}}$ is added to $\tilde{Y}_l$ after being processed via an optional $1 \times 1$ convolutional layer to obtain $\tilde{X}_{l-2} \in \mathbb{R}^{T\times m_{l}}$ to enable element-wise addition with $\tilde{Y}_l$, and then finally passed through a ReLU layer to obtain $X_l = ReLU(\tilde{Y}_l \EWA \tilde{X}_{l-2})$, where $\EWA$ is the element-wise addition operation.

\subsubsection{GAP layer to obtain fixed-dimensional vector for time series of varying lengths}
For classification tasks, a standard CNN approach would flatten the output of the last convolutional layer to obtain a $m_L \times T$-dimensional vector, and further use FC layer(s) before a final softmax layer. For long time series, i.e. large $T$, this approach leads to a significantly large number of trainable parameters that grows linearly with $T$. 
Instead, we pass the output of the final convolutional block $X_L \in \mathbb{R}^{T \times m_L}$ through a Global Average Pooling (GAP) layer that averages each feature map along the time dimension (as used in, e.g. \cite{wang2017time} and \cite{fawaz2018deep}). More specifically, GAP layer maps $X_L \in \mathbb{R}^{T \times m_L}$ to a vector $\mathbf{z} \in \mathbb{R}^{m_L}$ by taking a simple average of the $T$ values in each of the $m_L$ feature maps, thereby drastically reducing the number of trainable parameters. 

In a nutshell, CTN takes as input a univariate time series $x_{1\ldots T}$ of length $T$ and converts it to a fixed-dimensional feature vector $\mathbf{z}$ of length $m_L$ to be subsequently passed to a multi-head FC layer followed by softmax layer for training the various layers of CTN, as described next  and summarized in Algorithm \ref{algo}.

\subsection{Training CTN \label{ssec:training_ctn}}

\begin{figure}[h]
    \centering
    \includegraphics[scale=0.35,trim={2cm, 11.cm, 1cm, 6cm},clip]{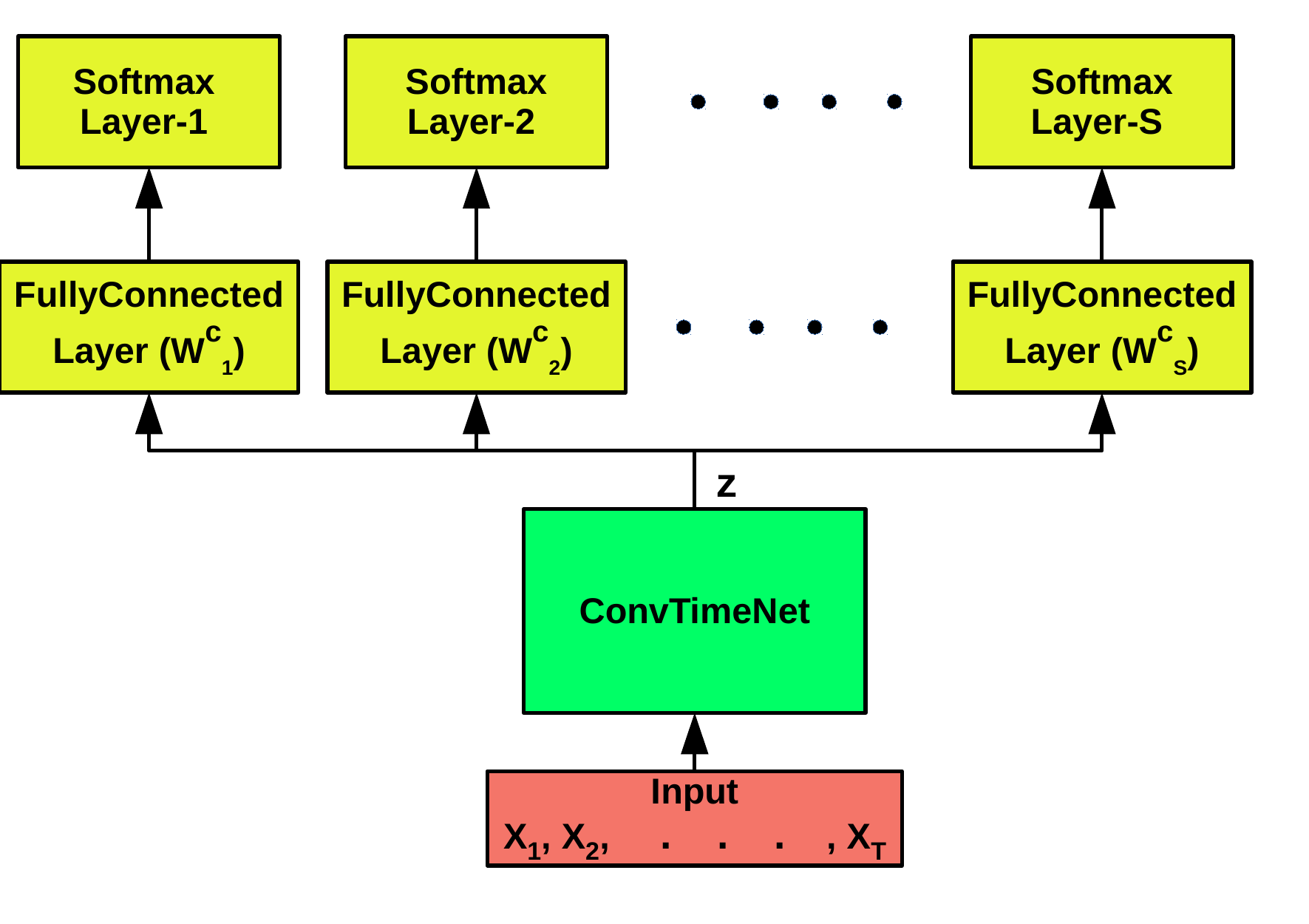}
    \caption{Training CTN using multi-head FC and softmax layer}
    \label{fig:training}
\end{figure}

Hereafter, we use $W_{CTN}$ to refer to the set of all trainable parameters of CTN consisting of $W_{k,l}$, $b_{k,l}$, and the BN parameters for $l=1,\ldots L$ and $k=1\ldots m_l$. 
In order to learn the parameters $W_{CTN}$, we train CTN over the diverse set of time series classification tasks in $\mathcal{S}$ with varying number of classes and time series lengths, by adopting a multi-head learning strategy: the core neural network (CTN) is common across the $S$ source tasks, while the task-specific parameters $W^c_{j}$ of the FC layer before the softmax layer are learned independently for each source task.
A labeled training dataset $\mathcal{D}_j \in \mathcal{S}$ ($j=1,\dots,S$) consists of $N_j$ samples and corresponds to a $K_j$-class classification problem. Since each dataset has different number of classes, we use $S$ FC and softmax layers, one for each dataset $\mathcal{D}_j$ as shown in Fig. \ref{fig:training}, with the $j$-th head mapping $\mathbf{z}$ to $K_j$ probability values.

Since the number of samples $N_j$ across datasets can vary significantly, we consider training CTN on $n$ randomly sampled batches of size $b$ for each of the $S$ datasets in an epoch. 
Each epoch, therefore, considers $n\times b$ training samples from each dataset $\mathcal{D}_j$ resulting in a total of $S\times n\times b$ training samples per epoch.
The order in which the $S$ datasets are iterated within an epoch is decided randomly. 
In turn, the $n$ batches of a dataset $D_j$ are processed together (one after the other) while updating $W_{CTN}$ and $W^c_j$ in each iteration using stochastic gradient descent to minimize cross-entropy loss:
\begin{equation}\label{eq:loss}
\mathcal{L}_j=-\sum_{i=1}^{b}\sum_{k=1}^{K_j}{y^{(i)}_klog(\hat{y}^{(i)}_k)}
\end{equation} 
where $\hat{y}^{(i)}_k$ is the probability that the $i$-th time series instance in batch belongs to class $k$, $y^{(i)}_k=1$ for the target class while $y^{(i)}_k=0$ otherwise.

Note that while the parameters $W_{CTN}$ are updated during each of the $S\times n$ iterations in an epoch, the task-specific parameters $W^c_j$ are updated only during the corresponding $n$ iterations of dataset $\mathcal{D}_j$ in an epoch and stored thereafter, until these are re-used and updated during the next epoch when processing the batches from $\mathcal{D}_j$. 
By using the same filters for all the $S$ UTSC tasks, the learned filters are likely to capture generic time series trends, patterns and features that are potentially useful for time series from other domains.

\subsection{Fine-tuning CTN for a target dataset\label{ssec:fine-tune}}

\begin{figure}[h]
	\centering
	\includegraphics[scale=0.35,trim={0cm, 17cm, 0cm, 4.5cm},clip]{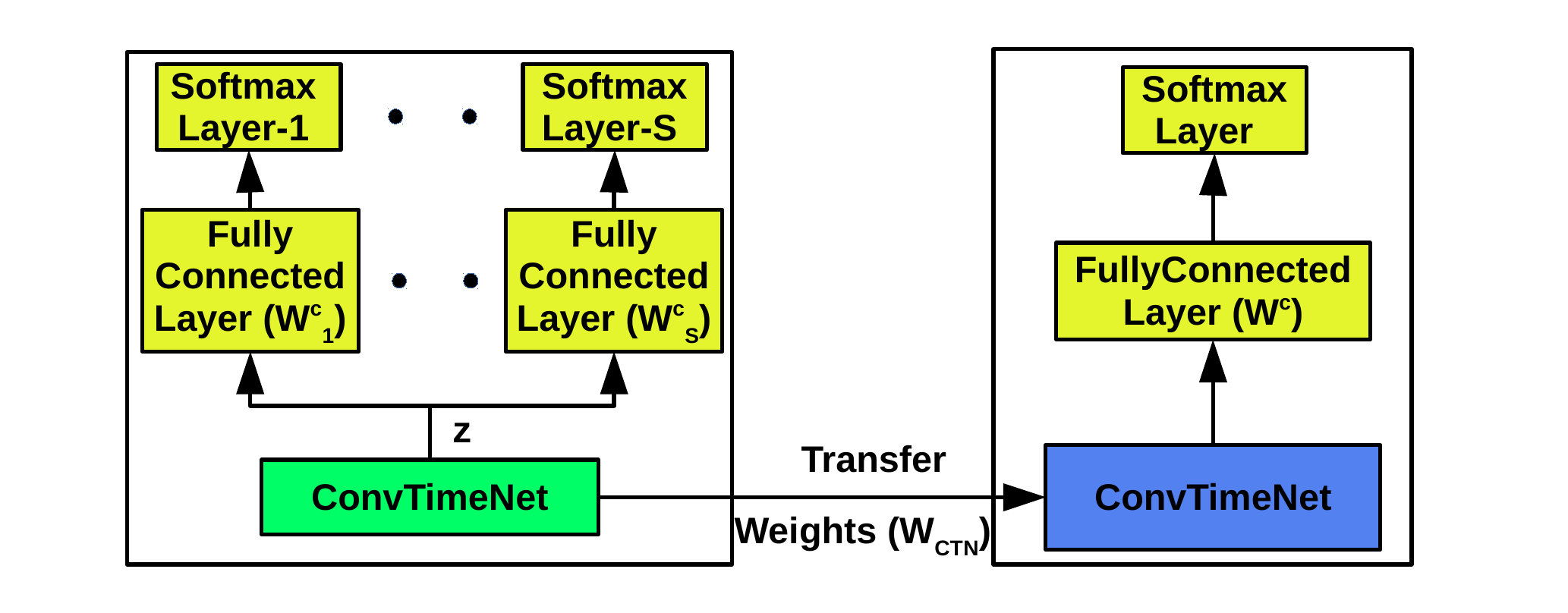}
	\caption{Transfer Learning using ConvTimeNet}
	\label{fig:tl_uts}
\end{figure}

\begin{algorithm}
	\footnotesize	
	\DontPrintSemicolon
	\SetAlgoLined
	\KwResult{Final CTN parameters $W_{CTN}$}
	\KwIn{Train set $\mathcal{S}$, validation set $\mathcal{S}^*$}
	\;
	Orthogonal Initialization of $W_{CTN,0}$\;
	\For{i=1 \ldots max\_train\_epochs}{
		$W_{CTN} \gets W_{CTN,i-1}$\;
		\tcp{train}
		\For{$\mathcal{D}_j \in \mathcal{S}$}{
			update $W_{CTN}$ and $W^c_j$ for $n$ iterations\; 
			}
		$W_{CTN,i} \gets W_{CTN}$\;

		\tcp{fine-tune for validation datasets}
		\For{$\mathcal{D}_k \in \mathcal{S}^*$}{
			$W_{i,k} \gets W_{CTN}$\;
			initialize $W^c_k$\;
			update $W_{i,k}$ and $W^c_k$ for $max\_val\_epochs$\;
			compute test loss $\mathcal{L}_{i,k}$ at epoch with minimum validation loss\;
		}
		$\mathcal{L}_i^v = \frac{1}{V}\sum_{\forall \mathcal{D}_k \in \mathcal{S}^*}{\mathcal{L}_{i,k}}$
	}
	$i^* = \argminA_{i} \mathcal{L}_i^v$\;
	$W_{CTN} \gets W_{CTN,i^*}$\;
	\;
\caption{Procedure for training CTN.}
\label{algo}
\end{algorithm}
We first describe how to fine-tune CTN for a new UTSC task, i.e. for a new dataset with a different set of target classes, and then discuss how to use it for (i) finding the best parameters $W_{CTN}$ via hold-out validation, as well as for (ii) transfer learning.
For a new UTSC task, we consider training the task-specific FC layer with parameters $W^c$ (followed by softmax) on top of the GAP layer of CTN while also updating the parameters $W_{CTN}$ as shown in Fig. \ref{fig:tl_uts}.
The parameters $W_{CTN}$ and $W^c$ are then updated together using cross-entropy loss function as in Equation \ref{eq:loss}.

\subsubsection{Fine-tuning for validation} 
Rather than the standard approach of using hold-out instances from the training dataset to build  a validation set, we use different datasets for validation, where we fine-tune CTN for hold-out unseen datasets independently one-at-a-time, and then use average loss across the validation datasets as validation loss (described later in this section). 
This way of validation of CTN mimics the transfer learning scenario where the goal is to adapt CTN for a new dataset, and therefore, yields a CTN model that is likely to generalize to unseen tasks. 
Such an approach for defining validation tasks has been shown to be useful in transfer learning settings, e.g. \cite{vinyals2016matching}. 

More specifically, for obtaining the best parameters $W_{CTN}$ during the iterative training process  (refer Algorithm \ref{algo}), we use a (relatively smaller) validation set $\mathcal{S}^*$ containing $V$ UTSC datasets such that $\mathcal{S}^* \cap \mathcal{S} = \emptyset$.
Let $W_{CTN,i}$ represent the parameters of CTN at the end of $i$-th training epoch.
The time series instances in $\mathcal{D}_k \in \mathcal{S}^*$ are divided into train, validate and test samples.
For each dataset $\mathcal{D}_k \in \mathcal{S}^*$ for $k=1,\dots,V$, the parameters $W_{CTN,i}$ and $W^c_k$ are fine-tuned using the train samples of $\mathcal{D}_k$ via stochastic gradient descent for a fixed number of epochs.
Using updated $W_{CTN,i}$ and $W^c_k$ at the epoch with minimum validation loss, we compute the test loss $\mathcal{L}_{i,k}$ for $\mathcal{D}_k$.  
Then, the validation loss for CTN at the end of $i$-th training epoch is defined as the average of these test losses across all datasets in $\mathcal{S}^*$, and is given by $\mathcal{L}_i^v=\frac{1}{V}\sum_{\forall \mathcal{D}_k \in \mathcal{S}^*}{\mathcal{L}_{i,k}}$. 
The optimal parameters $W_{CTN}$ are chosen at the epoch where the validation loss $\mathcal{L}_i^v$ is minimum, and represent the final parameters of the CTN. 

\subsubsection{Transfer to a new task}
The procedure for adapting / fine-tuning CTN for any new target task is similar to that of validation: We use $W_{CTN}$ as initial weights of CTN and randomly initialize FC layer weights $W^c$, and train them simultaneously for a fixed number of iterations using labeled data from target task.

\section{Experimental Evaluation\label{sec:exp}}
\begin{table*}[h]
	\centering
	\caption{Training and validation datasets used for ConvTimeNet. Here, T: time series length, C: number of classes, N: number of time series.} 
	\label{tab:datasets}
	{\subcaption{Training datasets}
		\begin{tabular}{|cccc|cccc|} \hline
			\textbf{Dataset} & \textbf{T} & \textbf{C} &\textbf{N}&\textbf{Dataset}& \textbf{T} & \textbf{C}&\textbf{N}\\ \hline
			\textbf{ItalyPowerDemand} & 24 & 2&1096 &\textbf{SonyAIBORobotSurfaceII} & 65 & 2&980\\    
			\textbf{SonyAIBORobotSurface} & 70 & 2&621 & \textbf{TwoLeadECG} & 82 & 2&1162\\
			\textbf{FacesUCR} & 131 & 14 &2250& \textbf{Plane} & 144 & 7&210\\
			\textbf{Gun\_Point} & 150 & 2 &200& \textbf{ArrowHead} & 251 & 3&211\\    
			\textbf{WordSynonyms} & 270 & 25 & 905& \textbf{ToeSegmentation1} & 277 & 2&268\\
			\textbf{Lightning7} & 319 & 7 & 143&\textbf{ToeSegmentation2} & 343 & 2&166\\
			\textbf{DiatomSizeReduction} & 345 & 4 &322& \textbf{OSULeaf} & 427 & 6&442\\
			\textbf{Ham} & 431 & 2 &214& \textbf{Fish} & 463 & 7&350\\
			\textbf{ShapeletSim} & 500 & 2&200 & \textbf{ShapesAll} & 512 & 60&1200\\
			\hline
		\end{tabular}
	}
	\centering
	{\subcaption{Validation datasets}
		\begin{tabular}{|cccc|cccc|} \hline      
			\textbf{Dataset} & \textbf{T} & \textbf{C}&\textbf{N} &\textbf{Dataset}& \textbf{T} & \textbf{C}&\textbf{N}\\ \hline
			\textbf{MoteStrain} & 84 & 2&1272 &\textbf{CBF} & 128 & 3&930 \\ 
			\textbf{Trace} & 275 & 4&200 & \textbf{Symbols} & 398 & 6 &1020\\ 
			\textbf{Herring} & 512 & 2&128 & \textbf{Earthquakes} & 512 & 2&461 \\\hline
		\end{tabular}
	}
	
\end{table*}

\begin{table*}[h]
	\scriptsize
	\centering
	\caption{Comparison of error rates of proposed CTN-T (transfer learning) with Flat COTE \cite{bagnall2015cote}, BOSS \cite{patrick2015boss}, ResNet \cite{fawaz2018deep}, and CTN-S (CTN-like architecture trained from scratch for target task).}
	\begin{tabular}{|p{2.0cm}|c|c|c|c|c||p{2.5cm}|c|c|c|c|c|}
		\hline
		\textbf{Dataset Name} & \multicolumn{1}{l|}{\textbf{\begin{tabular}[c]{@{}l@{}}Flat\\ COTE \end{tabular}}} & \multicolumn{1}{l|}{\textbf{BOSS}} &\multicolumn{1}{l|}{\textbf{ResNet} }&\multicolumn{1}{l|}{\textbf{\begin{tabular}[c]{@{}l@{}}CTN-S\end{tabular}}}&\multicolumn{1}{l||}{\textbf{\begin{tabular}[c]{@{}l@{}}CTN-T\\(proposed)\end{tabular}}} 
		&\textbf{Dataset Name} & \multicolumn{1}{l|}{\textbf{\begin{tabular}[c]{@{}l@{}}Flat\\ COTE \end{tabular}}} & \multicolumn{1}{l|}{\textbf{BOSS}} &\multicolumn{1}{l|}{\textbf{ResNet} }&\multicolumn{1}{l|}{\textbf{\begin{tabular}[c]{@{}l@{}}CTN-S\end{tabular}}}&\multicolumn{1}{l|}{\textbf{\begin{tabular}[c]{@{}l@{}}CTN-T\\(proposed)\end{tabular}}} \\ \hline
		\textbf{Adiac} & 0.21 & 0.24 &0.17 & 0.17 & \textbf{0.16}&\textbf{50words} & 0.20 & 0.29 & 0.26 & 0.17 & \textbf{0.16}\\
		\hline
		\textbf{Chlor.Conc.} & 0.27 & 0.34 & 0.16 & \textbf{0.14}& 0.17 &
		\textbf{Beef} & \textbf{0.13} & 0.20 & 0.25 & 0.31 & 0.26\\
		\hline
		\textbf{Cricket\_X} & 0.19 & 0.26 & 0.21 & \textbf{0.14} &
		\textbf{0.14} & \textbf{BeetleFly} & 0.20 & \textbf{0.10} & 0.15 &
		0.17 & 0.13\\
		\hline
		\textbf{Cricket\_Y} & 0.17 & 0.25 & 0.20 & \textbf{0.14} & \textbf{0.14} &
		\textbf{BirdChicken} & 0.10 & \textbf{0.05} & 0.11 & 0.18 & 0.17\\
		\hline
		\textbf{Cricket\_Z} & 0.19 & 0.25 & 0.19 & \textbf{0.12} & 0.13 &
		\textbf{Coffee} & \textbf{0.00} & \textbf{0.00} & \textbf{0.00} &\textbf{0.00} & \textbf{0.00} \\
		\hline
		\textbf{Dist.Phal.O.A.G} & 0.25 & 0.25 & 0.20 & 0.21 & \textbf{0.18} &
		\textbf{Dist.Phal.O.C} & 0.24 & 0.27 & \textbf{0.20} & 0.22 & 0.21\\
		\hline
		\textbf{Dist.Phal.TW} & 0.30 & 0.32 & \textbf{0.24} & 0.27 & 0.26 &
		\textbf{ECG5000} & \textbf{0.05} & 0.06 & 0.07 & 0.06 & 0.06 \\
		\hline
		\textbf{ECG200} & 0.12 & 0.13 & 0.12 & 0.14 &\textbf{0.08}&
		\textbf{ECGFiveDays} & \textbf{0.00} & \textbf{0.00} & 0.03 &
		\textbf{0.00} & \textbf{0.00}\\
		\hline 
		
		\textbf{ElectricDevices} & 0.29 & \textbf{0.20} & 0.27 & 0.29 & 0.30 & \textbf{FaceAll} & \textbf{0.08} &
		0.22 & 0.17 & 0.20 & 0.21\\
		\hline
		\textbf{FordA} & \textbf{0.04} & 0.07 & 0.08 & 0.05 & 0.06 &
		\textbf{FaceFour} & 0.10 & \textbf{0.00} & 0.05 & 0.05 & 0.03\\
		\hline
		\textbf{FordB} & 0.20 & 0.29 & 0.09 & \textbf{0.08} & \textbf{0.08}
		&\textbf{InsectWingbeatSound} & \textbf{0.35} & 0.48 & 0.49 & 0.36 &0.37\\
		\hline
		\textbf{Mid.Phal.O.A.G} & 0.36 & 0.46 & \textbf{0.27} & 0.29 & 0.28 &\textbf{MedicalImages} & 0.24 & 0.28 & 0.23 & 0.22 & \textbf{0.21} 
		\\
		\hline
		\textbf{Mid.Phal.O.C} & 0.20 & 0.22 & \textbf{0.19} & \textbf{0.19} & \textbf{0.19} &
		\textbf{Mid.Phal.TW} & 0.43 & 0.46 & 0.40 & 0.41 & \textbf{0.39} \\
		\hline
		\textbf{PhalangesO.C} & 0.23 & 0.23 & \textbf{0.16} & 0.17 & 0.17 & \textbf{Meat} & 0.08 & 0.10 & \textbf{0.03} & 0.11 &
		0.09\\
		\hline
		\textbf{Prox.Phal.O.A.G} & \textbf{0.15} & 0.17 & \textbf{0.15} &
		0.16 & 0.16&
		\textbf{Prox.Phal.O.C} & 0.13 & 0.15 & \textbf{0.08} & 0.10 & 0.09 \\
		\hline
		\textbf{Prox.Phal.TW} & 0.22 & \textbf{0.20} & 0.21 & 0.22 &
		 0.22 &
		\textbf{Strawberry} & 0.05 & \textbf{0.02} & 0.04 & 0.03 &
		0.03 \\
		\hline
		\textbf{SwedishLeaf} & 0.05 & 0.08 & \textbf{0.04} & \textbf{0.04} &
		\textbf{0.04} & 
		\textbf{synthetic\_control} & \textbf{0.00} & 0.03 & \textbf{0.00} &\textbf{0.00} & \textbf{0.00}\\
		\hline
		\textbf{Two\_Patterns} & \textbf{0.00} & 0.01 & \textbf{0.00} &
		\textbf{0.00} & \textbf{0.00} & 
		\textbf{uWave\_X} & 0.18 & 0.24 & 0.22 &\textbf{0.17} & \textbf{0.17} \\
		\hline
		\textbf{uWave\_Y} & 0.24 & 0.32 & 0.33 & 0.24 & \textbf{0.23} &
		\textbf{uWave\_Z} & 0.25 & 0.31 & 0.25 & \textbf{0.23} & \textbf{0.23} \\
		\hline
		\textbf{wafer} & \textbf{0.00} & 0.01 & \textbf{0.00} &\textbf{0.00} & \textbf{0.00}&\textbf{Wine} & 0.35 & 0.26 & 0.26 & \textbf{0.17} & \textbf{0.17}
		\\ \hline
		\textbf{yoga} & 0.12 & \textbf{0.08} & 0.13 & 0.10 & \textbf{0.08} 
		&\textbf{W/T/L of CTN-T} & 26/6/9 & 30/4/7 &
		22/6/13 & 17/18/6 & -\\
		\cline{1-6}
		\multicolumn{6}{c|}{}& \textbf{Mean Arithmetic Rank} & 3.22 & 3.91 & 2.88 &
		2.72 & \textbf{2.27}\\
		\cline{7-12}
	\end{tabular}
	\label{table:comparison}
\end{table*}

We empirically evaluate CTN from three perspectives: 
	1) \textit{classification performance}: to evaluate if fine-tuning CTN for target task provides better accuracy compared to training a model from scratch,
	2) \textit{computational efficiency}: to evaluate if CTN can be adapted quickly with fewer iterations compared to training a deep model from scratch,
	3) \textit{ablation study}: to understand the advantage of multiple filter lengths in CTN.
Additionally, we provide a qualitative analysis of the trained filters in CTN and useful insights into the interpretability of results in Sec \ref{ssec:filter-analysis}.

\subsection{Dataset details}

We train and test CTN on diverse disjoint subsets of the $85$ datasets taken from the UCR TSC Archive Benchmark \cite{UCRArchive,bagnall2017great,fawaz2018deep} belonging to seven diverse categories: Image Outline, Sensor Readings, Motion Capture, Spectrographs, ECG, Electric Devices and Simulated Data.
All time series are z-normalized, i.e. the mean and standard deviation of the values in any time series is 0 and 1, respectively.
The length of time series varies significantly from $24-2709$ across datasets, and the number of classes also varies significantly from $2-60$. Further, the number of labeled training instances varies between $16-8926$.
We use the same (random) split of training ($\mathcal{S}$) and validation ($\mathcal{S}^*$) datasets (refer Table \ref{tab:datasets}) as used in \cite{timenet} and detailed in \cite{timenetarxiv}, such that we have $S=18$ datasets for training CTN and $V=6$ for model selection. 
The 18 training and 6 validation datasets have $T\leq 512$, and we therefore, restrict testing to the remaining 41 datasets with $T\leq 512$. 

For each training dataset, all the labeled train as well as test samples from the original train-test split in the archive are used for training CTN.
For each of the 6 validation datasets and the 41 test datasets, we use the same train-test splits as provided in \cite{bagnall2017great} while fine-tuning CTN using train split of respective datasets.

\subsection{Hyperparameters}
Based on preliminary experiments on a smaller subset of $\mathcal{S}$ and $\mathcal{S}^*$ to decide the number of layers and filters, we consider CTN with $L=4$ blocks, each convolutional layer in the convolutional blocks consist of five different filters lengths, i.e. $f\in\{4,8,16,32,64\}$, with 33 filters for each $f$ such that $m_l=33\times 5 = 165$  for each convolutional block ($l= 1,2,3,4$). We use the Adam optimizer for optimizing the weights of the networks with an initial learning rate of $0.002$.
We used orthogonal initialization of convolutional filters in all our experiments.
CTN was trained for $200$ epochs; during each epoch, for each dataset $\mathcal{D}_j \in \mathcal{S}$, we randomly chose $n=5$ batches of size $b=16$ each. For validation datasets, we fine-tune CTN parameters and task-specific parameters for $50$ epochs with a learning rate of $0.002$.
While adapting CTN for each test dataset, parameters of CTN and the FC layer are fine-tuned for $12000$ iterations with a reduced learning rate of $2\times 10^{-4}$. 
\subsection{Baselines considered}
We refer to the proposed approach of fine-tuning the pre-trained CTN for target task as \textbf{CTN-T} (CTN-Transfer), and compare it to:
(i) \textbf{CTN-S}: We train CTN-S (CTN architecture trained from Scratch) as an exact replica of CTN with all parameters initialized randomly for each test dataset. By doing so, we can attribute the gains in performance, if any, obtained via using CTN-T over CTN-S to the pre-trained filters in CTN. 
(ii) \textbf{ResNet} \cite{fawaz2018deep} as the state-of-the-art deep learning approach:
	ResNet is trained independently for each dataset and contains 11 layers of which the first 9 layers are convolutional with shortcut residual connections between residual blocks (each block with 3 convolutional layers) followed by a GAP layer and a FC layer+softmax. 
(iii) Two non-deep learning state-of-the-art techniques as baselines \cite{bagnall2017great}: i) Flat \textbf{COTE} (Collective of transformation-based ensembles) \cite{bagnall2015cote}, ii) \textbf{BOSS} (Bag of SFA Symbols)\cite{patrick2015boss}.

For evaluating CTN-T and CTN-S on each test dataset, we use the entire train split for training and use the model parameters corresponding to the iteration with minimum training cross-entropy loss, following the same protocol\footnote{We additionally considered a stratified sampling approach to divide the train split of each dataset into 75\%-25\% training and validation samples, and still found the resulting variant of CTN-T to perform better than BOSS, ResNet and Flat COTE methods used for comparison in Table \ref{table:comparison}. However, this variant was worse compared to the CTN-T model using entire train split for training; especially for three datasets, namely Beef, Chlor.Conc., and 50words. This can be attributed to a small number of training instances per class and/or large diversity in patterns within samples of same class.} as used in \cite{wang2017time,fawaz2018deep}. We train three models for each dataset (with randomly initialized FC layer and entire network for CTN-T and CTN-S, respectively), and report the average of the three error rates in Table \ref{table:comparison}.

\subsection{Observations}
\begin{figure}
		\begin{subfigure}[b]{0.45\columnwidth}
			\centering
			\includegraphics[scale=0.3,trim={0cm, 0cm, 0cm, 0cm},clip]{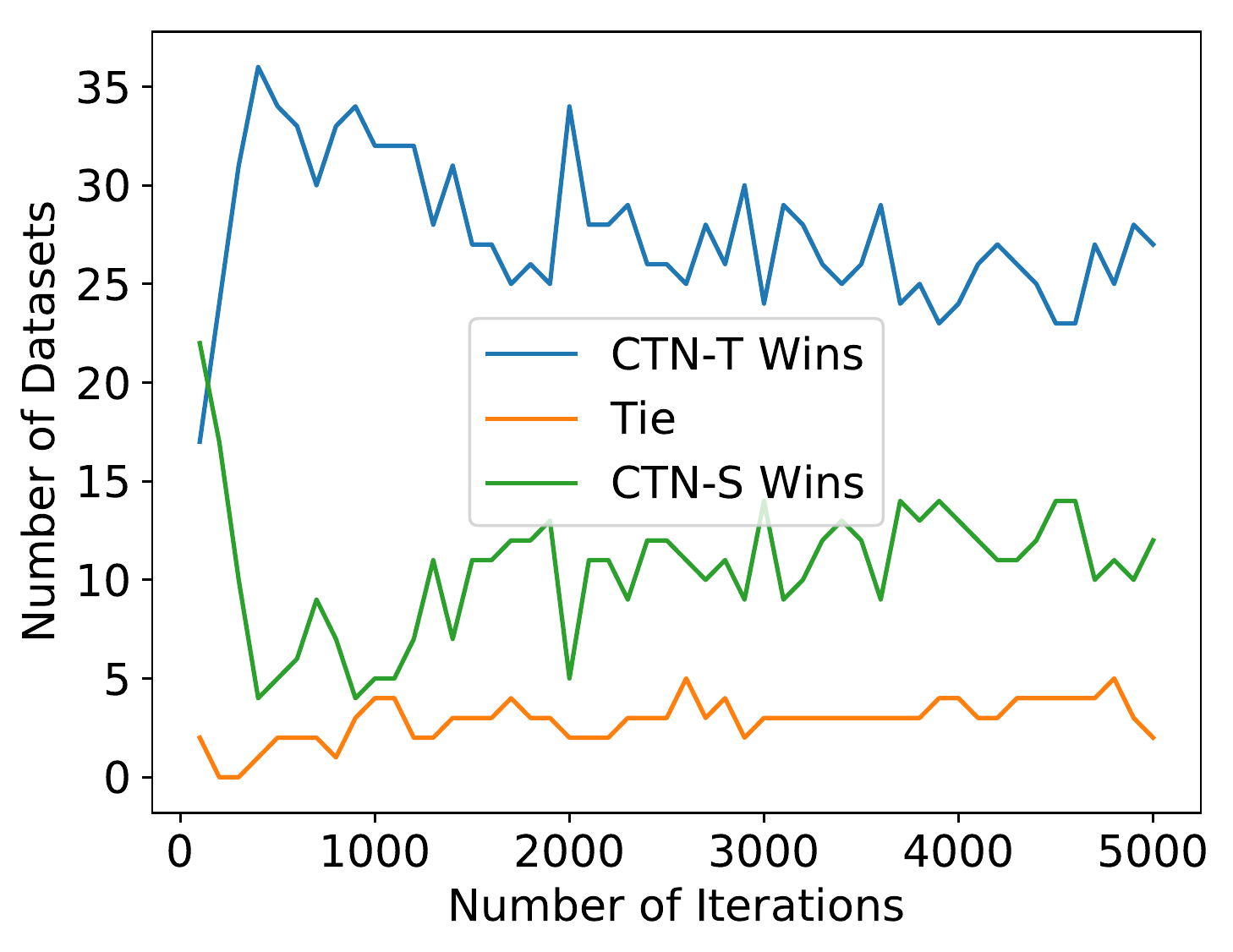}
			\caption{CTN-T vs. CTN-S}
			\label{fig:err-vs-ep-ctn-s}
		\end{subfigure}
~
		\begin{subfigure}[b]{0.45\columnwidth}
				\centering
				\includegraphics[scale=0.3]{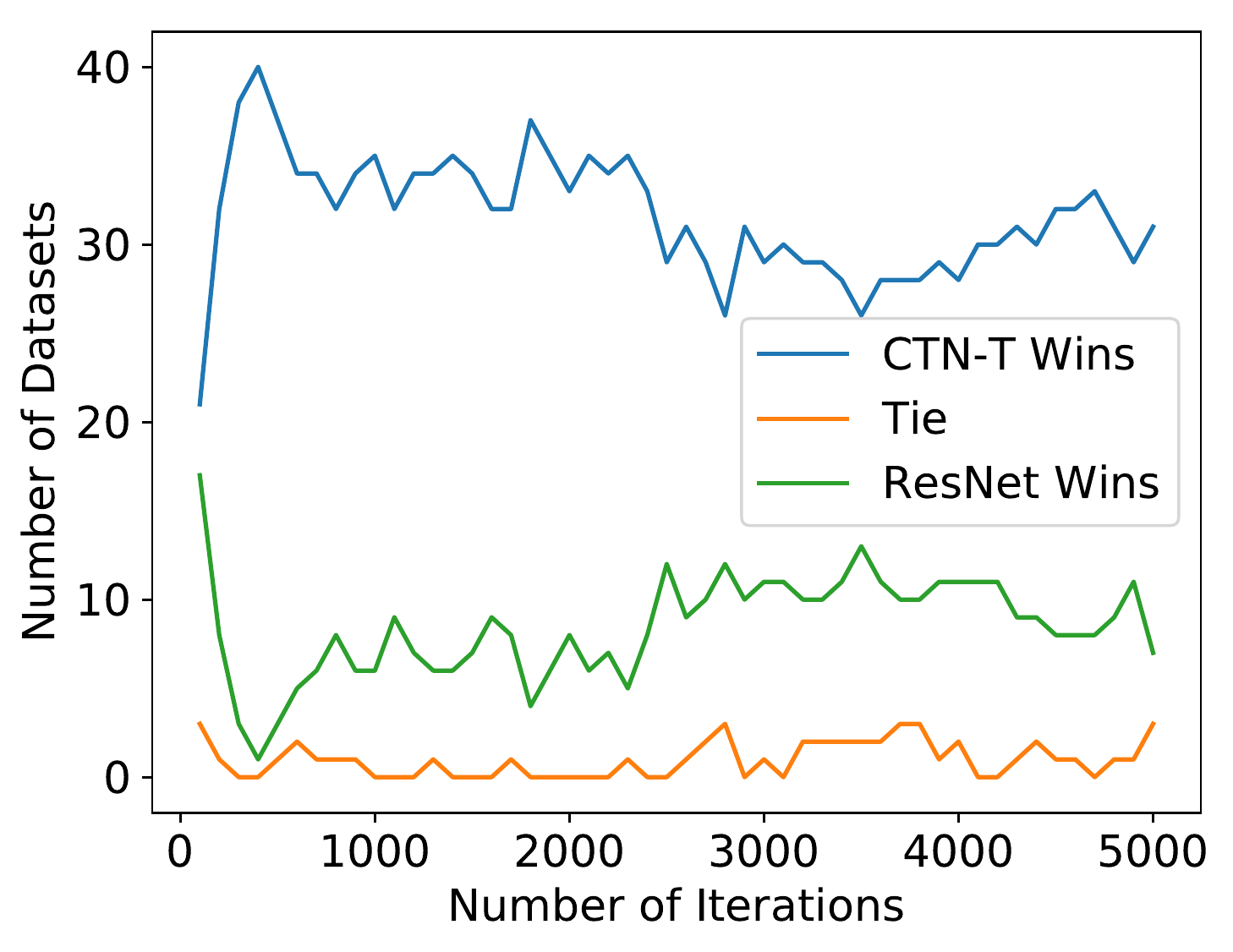}
				\caption{CTN-T vs. ResNet}
				\label{fig:err-vs-ep-resnet}
		\end{subfigure}
			
		\caption{Comparison in terms of no. of wins with no. of training iterations. CTN-T with pre-trained filters adapts faster to target tasks compared to models trained from scratch.}
		\label{fig:err-vs-epoch}
\end{figure}

The comparison of classification error rates (fraction of wrongly classified instances) and the number of wins/ties/losses (W/T/L) is summarized in Table \ref{table:comparison}.
We make the following key observations:
\begin{enumerate}[wide, labelwidth=!, labelindent=0pt]	
	\item CTN-T has W/T/L of 17/18/6 concluding that a pre-trained network based transfer learning (CTN-T) has significantly better performance compared to training the CTN-like architecture from scratch (CTN-S), as also highlighted in Fig. \ref{fig:ctn-t-vs-ctn-s}. 
	Further, we observe that CTN-T has W/T/L of 22/6/13 compared to ResNet, proving the advantage of leveraging a pre-trained model. CTN-T has mean arithmetic rank of 2.27 based on error rates which is significantly better than both non-transfer-based deep learning approaches, i.e. CTN-S and ResNet.
	\begin{figure}[h]
		\begin{subfigure}[b]{0.48\columnwidth}
			\centering
			\includegraphics[scale=0.28]{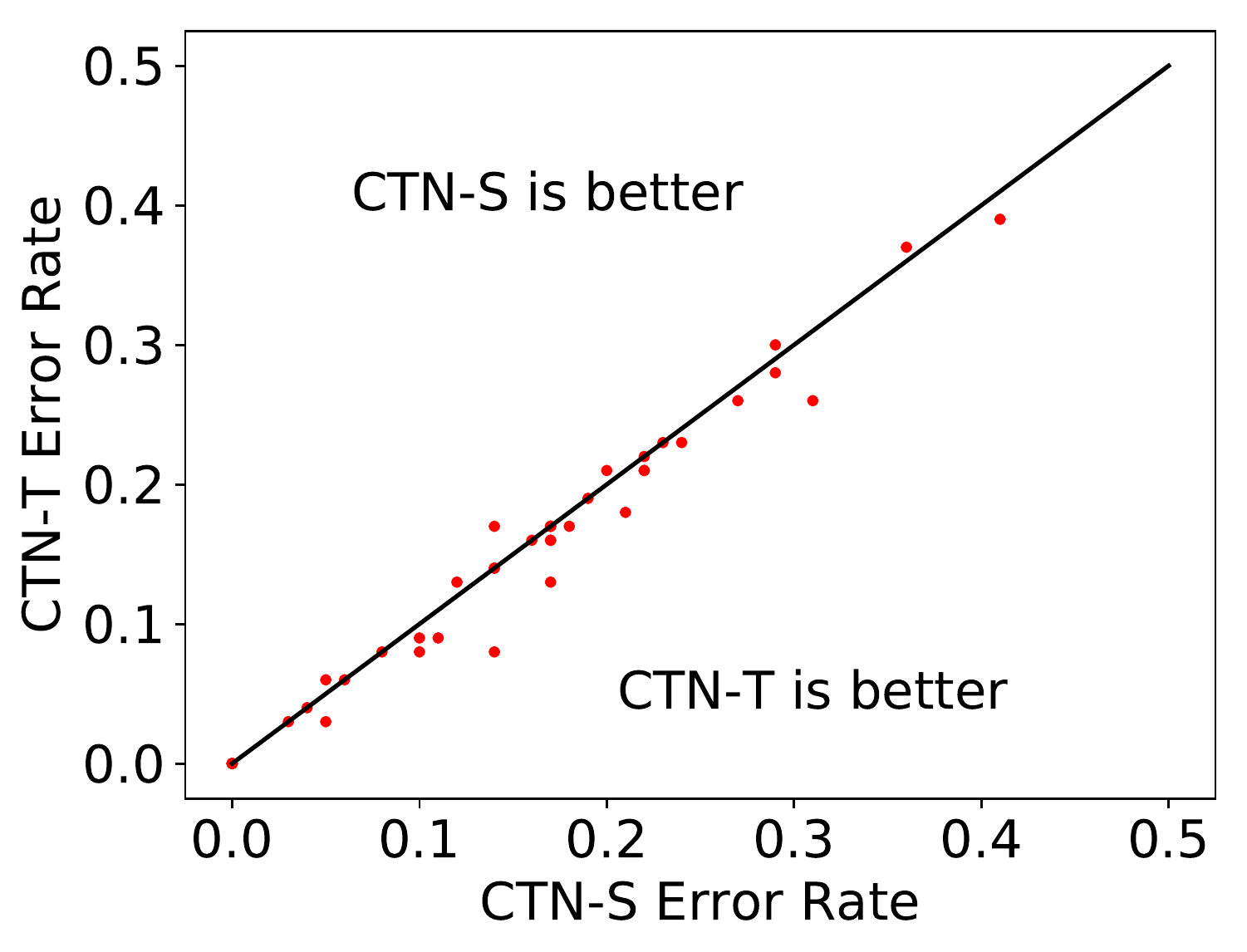}
			\caption{CTN-T (transfer learning) vs. CTN-S (training from scratch)}
			\label{fig:ctn-t-vs-ctn-s}
		\end{subfigure}
		~
		\begin{subfigure}[b]{0.48\columnwidth}
			\centering
			\includegraphics[scale=0.28]{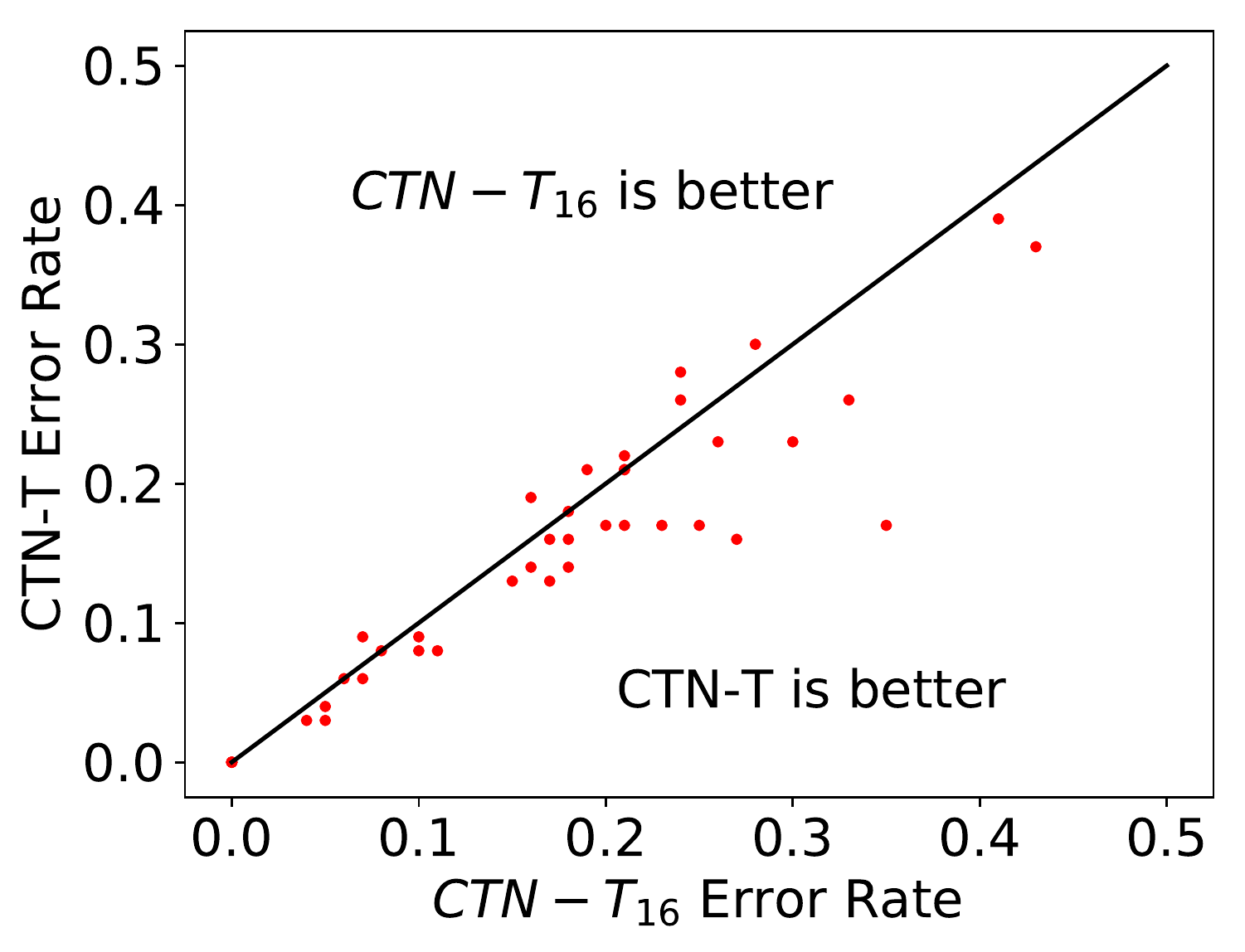}
			\caption{CTN-T ($f=4,8,16,32,64$) vs. CTN-T$_{16}$ ($f=16$).}
			\label{fig:ctn-t-vs-ctn-32}
		\end{subfigure}
		\caption{Scatter plots of classification error rates.}
	\end{figure}
	
	\item As shown in Fig. \ref{fig:err-vs-epoch}, we observe that CTN-T performs significantly better compared to CTN-S and ResNet when number of parameter updates is small, i.e. fewer number of training/fine-tuning iterations. (Due to random initialization of FC layer, ResNet and CTN-T are similar initially but CTN-T quickly adapts.) This suggests that starting from a pre-trained model is computationally efficient compared to starting from scratch: CTN-T takes fewer iterations to reach optimal classification performance while having better classification error rates, proving the advantage of leveraging a pre-trained network over a network trained from scratch.
	
	\item CTN-T has W/T/L of 26/6/9 compared to COTE. Given that COTE is extremely computationally expensive \cite{bagnall2017great}, training and deploying it in practical applications can be highly inefficient. On the other hand, training and inference in CTN-T is highly parallelizable making it suitable for practical applications.
	Further, fine-tuning of pre-trained CTN is efficient and overcomes the need for any hyperparameter tuning.
\end{enumerate}

\subsection{Ablation study: Does having different filter lengths help? \label{ssec:ablation}}
To evaluate the importance of having multiple filter lengths in a transfer learning setting to deal with diverse datasets, we train four CTN-like architectures keeping the filter length $f$ fixed ($8$, $16$, $32$, and $64$) for all layers while keeping the total number of trainable parameters to be approximately same as that in CTN by suitably adjusting the number of filters in each layer, such that we have $290$, $200$, $145$ and $100$ filters when $f=8$, $16$, $32$, and $64$, respectively, in each convolutional layer. 
We observe that CTN-T performs significantly better in comparison to any of these variants. CTN-T has W/T/L of 24/10/7 compared to the best performing fixed-length variant CTN-T$_{16}$ corresponding to $f=16$ as shown in Fig. \ref{fig:ctn-t-vs-ctn-32}. These results highlight the significance of having filters of multiple lengths in a transfer learning setting: multiple filter lengths help to capture trends and patterns occurring at varying temporal resolutions which would be otherwise difficult to capture via filters of fixed length in a CNN model.

\subsection{Analysis of CTN filters\label{ssec:filter-analysis}}
\begin{figure*}[h]
	\centering
	\begin{subfigure}[b]{0.6\columnwidth}
		\centering
		\includegraphics[scale=0.33,trim={0cm, 0cm, 0cm, 0cm},clip]{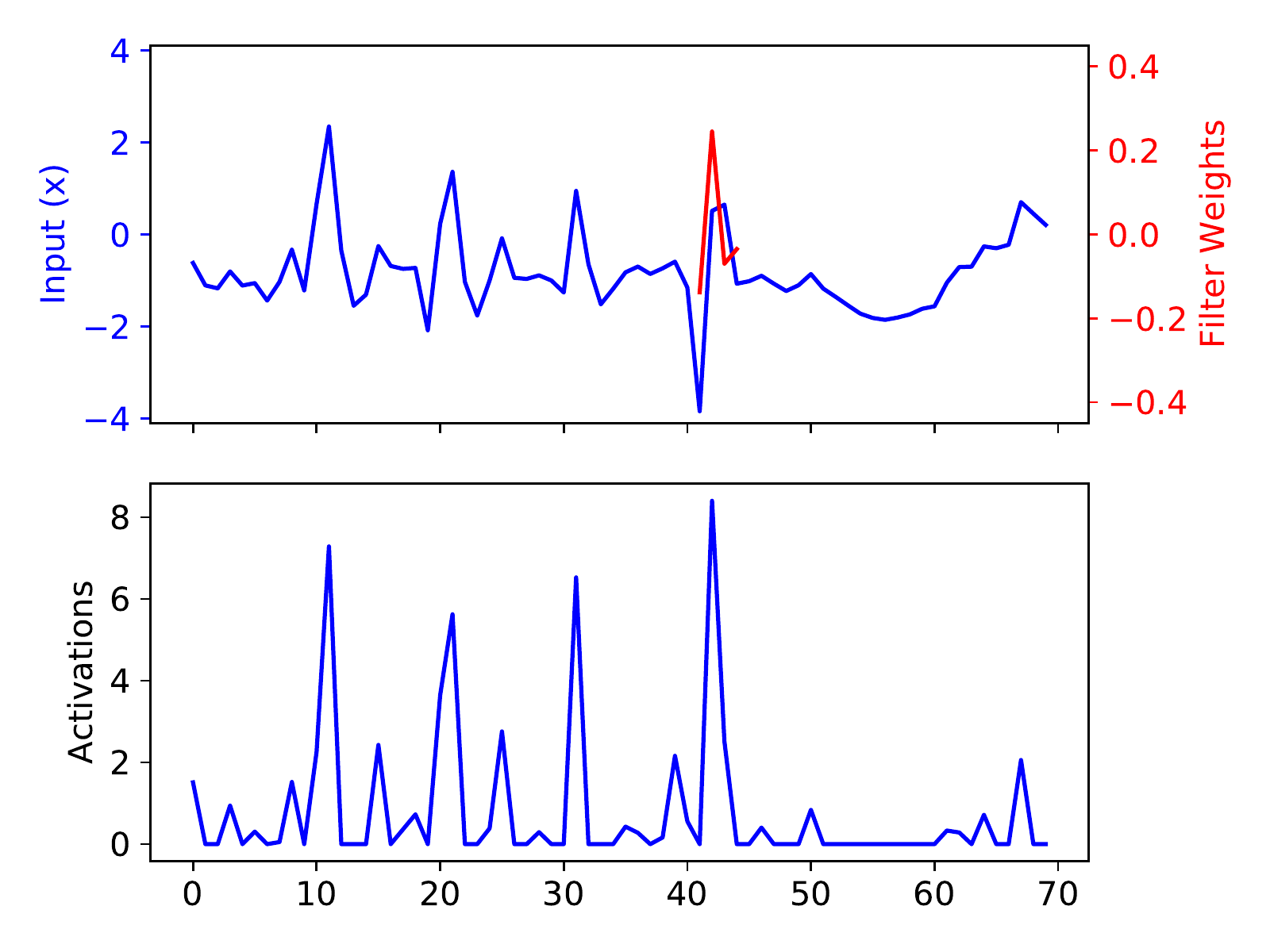} 
		\caption{Sharp rise \& fall (Cricket-X, $f=4$)}
		\label{fig:act1}
	\end{subfigure}
	\begin{subfigure}[b]{0.6\columnwidth}
		\centering
		\includegraphics[scale=0.33,trim={0cm, 0cm, 0cm, 0cm},clip]{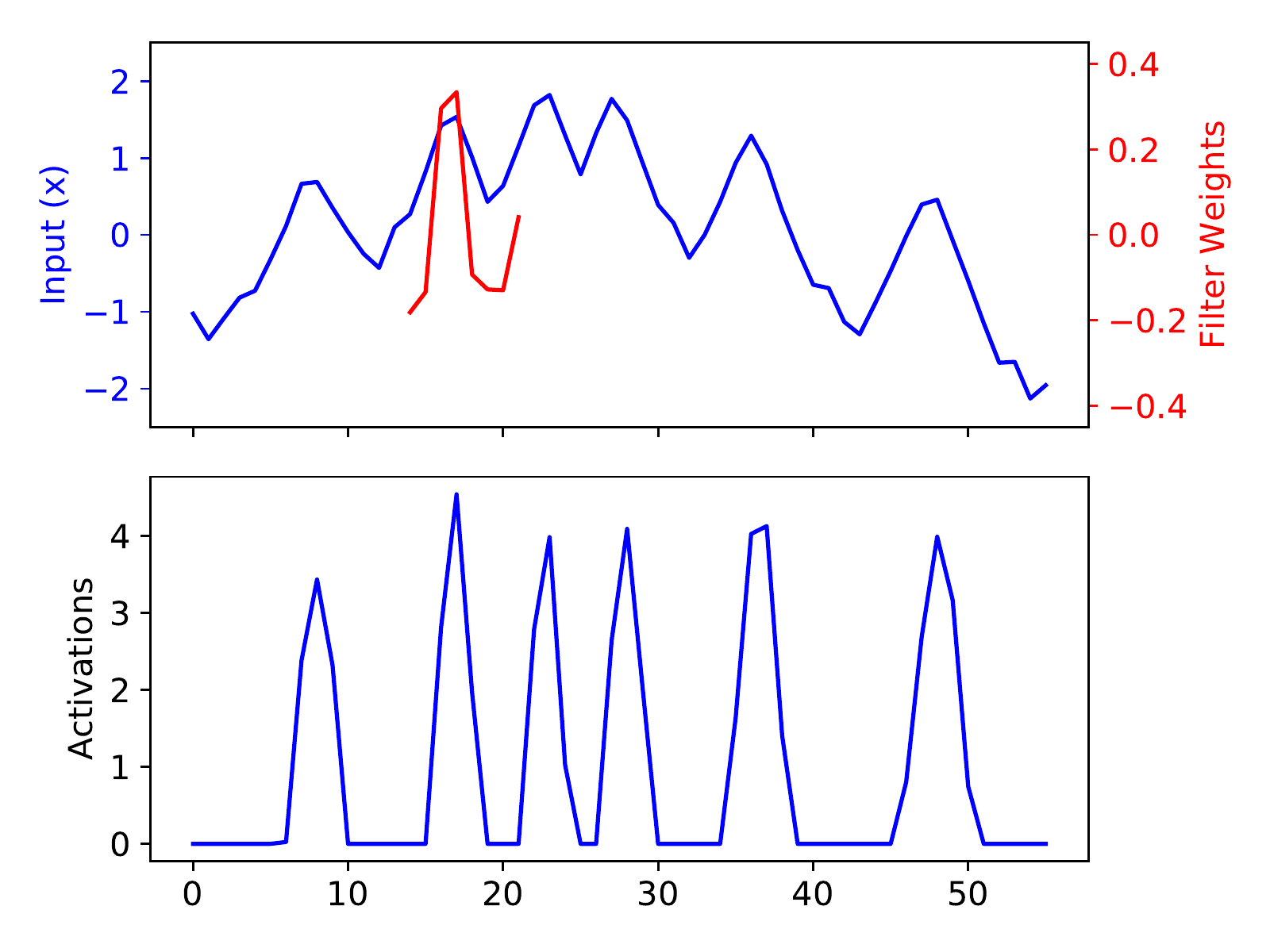}
		\caption{Gradual rise \& fall (Swe.Leaf, $f=8$)}
		\label{fig:act2}
	\end{subfigure}
	\begin{subfigure}[b]{0.6\columnwidth}
		\centering
		\includegraphics[scale=0.33,trim={0cm, 0cm, 0cm, 0cm},clip]{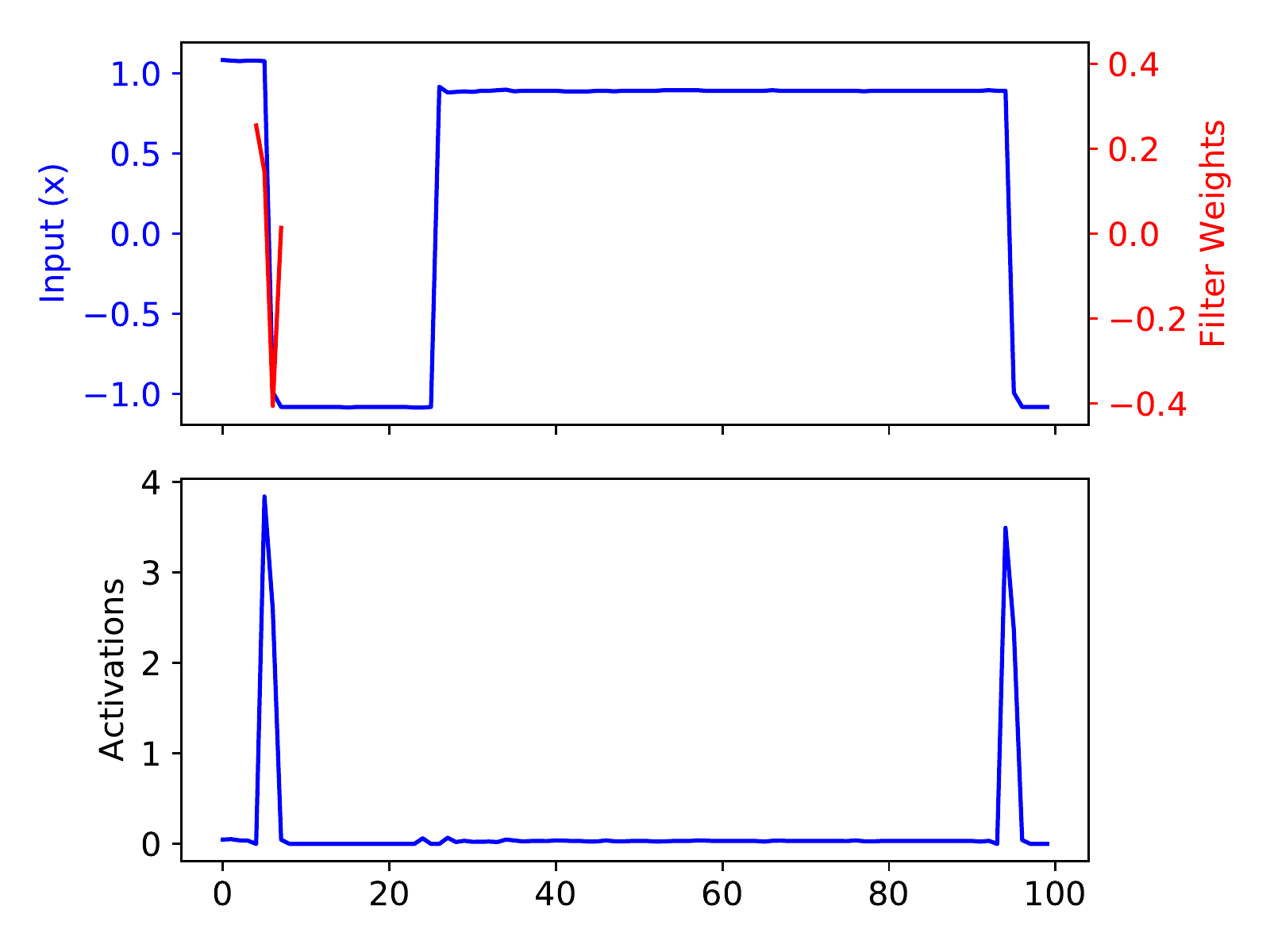}
		\caption{Sharp fall (Wafer, $f=4$)}
		\label{fig:act4}
	\end{subfigure}
	\caption{Filters capturing different basic time series patterns: Sample time series (input) with most relevant filter from 1st conv. layer (filter weights) in CTN-T, and feature maps (activation). Filter is shown at the point where activation is maximum.  }
	\label{fig:act}
\end{figure*}
\subsubsection{Fine-tuning all layers vs partial fine-tuning}
Typically, lower layers of a deep neural network tend to learn generic features while higher layers tend to learn task-specific features.
To analyze this behavior in CTN, we consider four variants where we freeze the parameters of the first, first two, first three, and all four convolutional layers of CTN, respectively, while fine-tuning the remaining layers for a test dataset as described in Section \ref{ssec:fine-tune}. 
We keep the parameters of BN layers trainable and only freeze the convolutional layers for reasons explained in \cite{carl2017multi}.
We observe an average improvement in classification performance of around 1\% across the test datasets by freezing the first convolutional block, a drop of around 0.9\% when freezing first two or three layers, while a significant drop of $3\%$ on freezing all four convolutional layers. 
These observations suggest that fine-tuning the final convolutional layer can be critical to obtain good task-specific models from pre-trained CTN.
Further, minor improvement by freezing the first layer can be attributed to the fact that it may be capturing generic patterns relevant across datasets. 

\subsubsection{Qualitative analysis of filters from first layer of ConvTimeNet} 
\begin{figure}[h]
 	\centering
 	\includegraphics[scale=0.2]{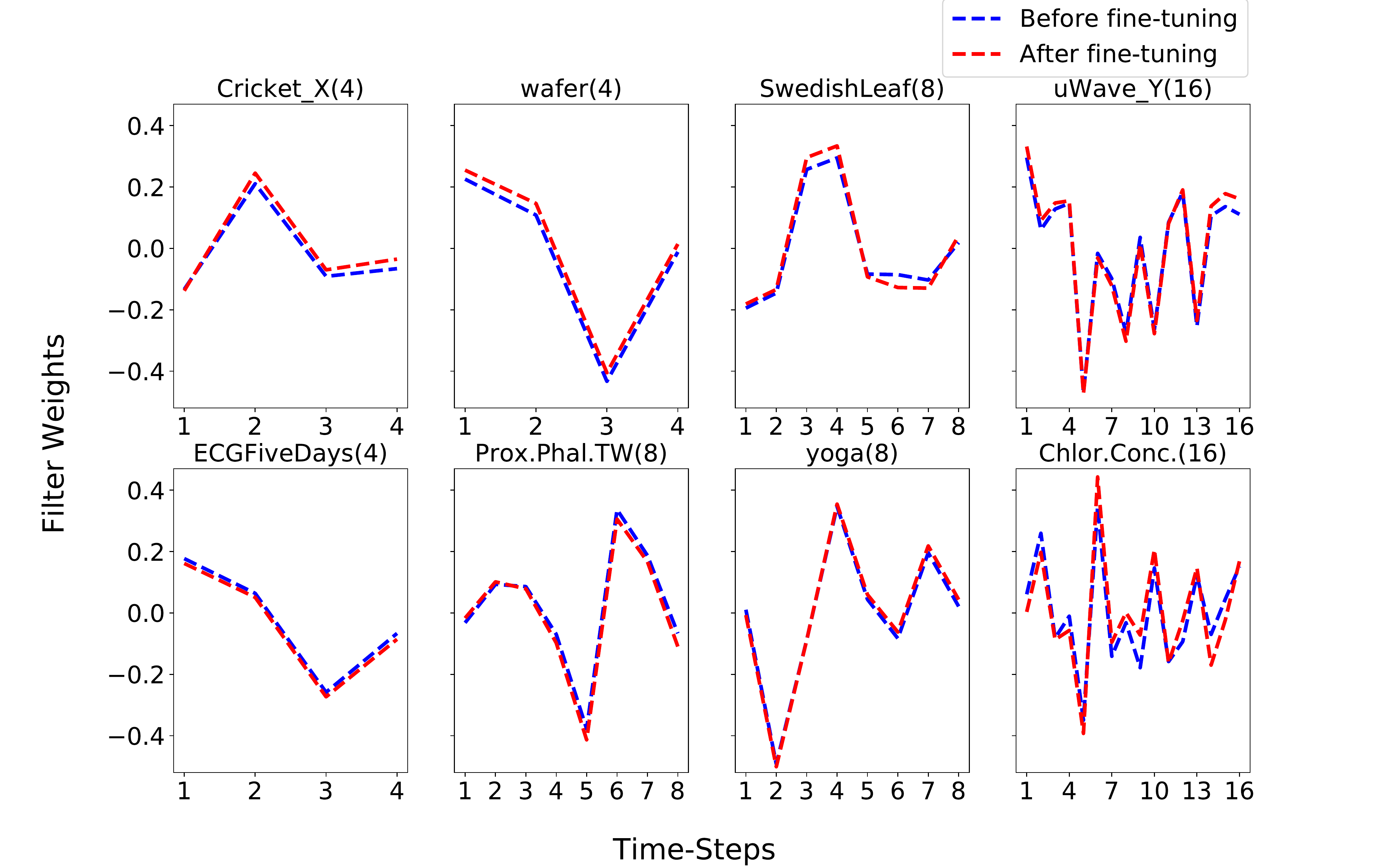}
 	\caption{Most relevant filters from first layer in CTN for sample datasets. Numbers in bracket represent the filter length $f$.}
 	\label{fig:filters}
 \end{figure}

We first find the filter with maximum value for \textit{relevance} $r(W_{k,1})$ for a dataset, where $k = 1\ldots 165$, and $W_{k,1} \in \mathbb{R}^f$ with $f \in \{4,8,16,32,64\}$:
\begin{equation}
\label{eq:relevance}
 r(W_{k,1}) = \frac{1}{N}\sum_{i=1}^ N max(W_{k,1}*x^{(i)}_{1\ldots T} + b_{k,1})
\end{equation} 
Fig. \ref{fig:filters} depicts the filter weights for eight different test datasets before and after fine-tuning of CTN.
We observe that the filters capture typical patterns that are encountered in time series like sharp/gradual rise/fall, rise followed by a fall, etc. further indicating generic features learned by CTN which do not change much on fine-tuning. We found different filters to be most relevant for different datasets. The patterns captured are illustrated in Fig. \ref{fig:act} using filter weights and corresponding activations for sample time series. 
Further, it is interesting to note that some of the filters (e.g. the ones with $f=16$ in Fig. \ref{fig:filters}) may appear extremely noisy and not capturing any trends at first. However, ignoring those points in the filters with weights very close to $0$ (say, between $-0.05$ to $0.05$) yields meaningful patterns - these filters tend to capture trends over longer time steps while ignoring some of the steps (corresponding to filter weight $=0$) in a time series: e.g. refer the most relevant filter for uWave\_Y dataset with $f=16$ which is $\approx 0$ at time steps $6$ and $8$, i.e. this filter ignores the $6$th and $8$th time steps in a window of length $16$ during the convolution operation, and therefore tries to capture coarser higher level temporal patterns rather than finer trends.
\subsubsection{Interpretability via Occlusion Sensitivity}
\begin{figure}[h]
    \centering
    \includegraphics[width=0.5\columnwidth,height=0.6\columnwidth]{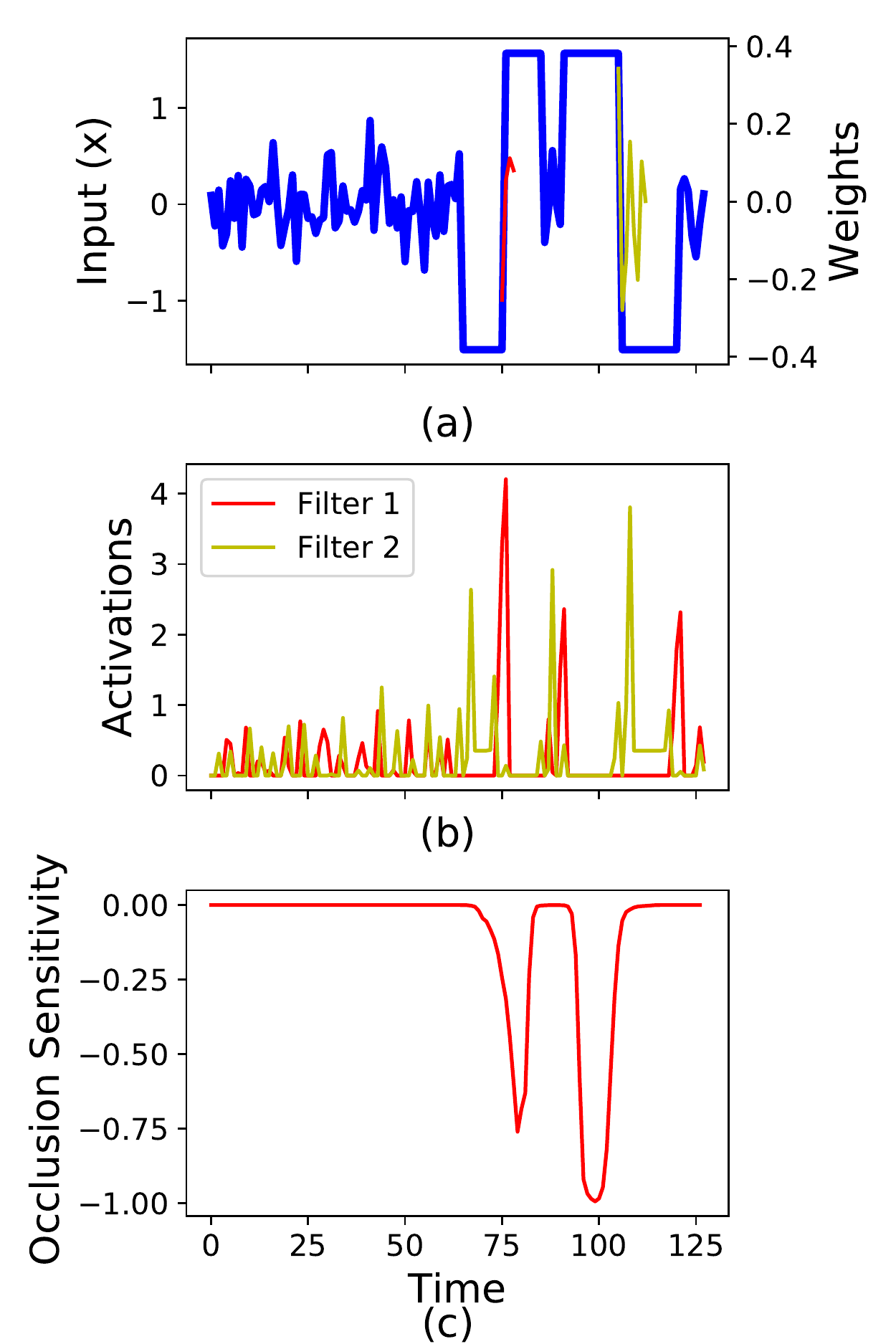}
    \caption{(a) Sample time series with top-2 relevant filters from Two Patterns dataset, (b) their activation maps, and (c) occlusion sensitivity plot.}
    \label{fig:interpret}
\end{figure}

We provide preliminary analysis of interpretability in terms of identifying the region(s) in the time series that are most relevant for making a particular classification decision. We use the ``Two Patterns" test dataset from ``Simulated" category as an illustrative example for its ease of visual interpretability: 
``Two Patterns" has four classes constituting the possible combinations of the two patterns ``up" and ``down". Refer Fig. \ref{fig:interpret}(a) for a (test) instance of up-down class along with the two most relevant filters for this dataset identified using Eq. \ref{eq:relevance}. 
We observe that one filter captures the ``up" trend while the other captures the ``down" trend, with maximum activation value coming at the corresponding points in the time series as depicted in Fig. \ref{fig:interpret}(b). 
To find the region of time series used by CTN-T classifier to arrive at the classification decision, we compute occlusion sensitivities \cite{zeiler2014visualizing} by occluding parts of the time series and observing the changes in probability $\hat{y}$ for the predicted class. Specifically, we consider a moving window of length $0.1T$ and set the values over that window to $0$. 
The moment an important part of the time series is occluded, we expect a sharp drop in the probability $\hat{y}$ for the predicted class. 
This change in probability over time, i.e. occlusion sensitivity $s_t = \hat{y}^o_t - \hat{y}^a_t$ (where $\hat{y}^o_t$ is probability for the predicted class after occluding and $\hat{y}^a_t$ is the actual probability without any occluded parts) is shown in Fig. \ref{fig:interpret}(c). We observe that as soon as the window covers the ``up"/``down" trend in the time series, there is a sharp drop in $s_t$ indicating that the network is focusing on the correct regions in the time series for making the decisions (also these regions coincide with the most relevant filters for the dataset). 

\section{Conclusion and Future Work\label{sec:conc}}
We have proposed ConvTimeNet (CTN): a pre-trained deep CNN for univariate time series classification. 
CTN leverages multiple length filters to model various temporal patterns from diverse time series across domains. 
Adapting a pre-trained model like CTN for the target task via fine-tuning i) yields significantly better results compared to existing state-of-the-art time series classification approaches, ii) is computationally efficient, and iii) does not require expertise in deep learning compared to training a deep network from scratch. 
In future, we plan to train a bigger CTN model on a larger and diverse dataset with longer time series.
Also, it will be interesting to see if the number of parameters to be updated during the fine-tuning task can be reduced to make fine-tuning even more efficient.

\bibliographystyle{IEEEtran}
\bibliography{BibTex/sensor_analytics,BibTex/esann17,BibTex/ijcnn19}


\end{document}